\documentclass[11pt]{article}
\usepackage{eamt24}
\usepackage{times}
\usepackage{url}
\usepackage{latexsym}
\usepackage{booktabs}
\usepackage[small,bf]{caption} 
\usepackage{longtable}
\title{Enhancing Scientific Discourse: Machine Translation for the Scientific Domain}

\author{Dimitris Roussis, Sokratis Sofianopoulos, Stelios Piperidis\\
\\
Institute for Speech and Language Processing, Athena RC\\ Artemidos 6 \& Epidavrou, Athens, Greece\\
{\tt \{dimitris.roussis, s\_sofian, spip\}@athenarc.gr}}

\begin{document}
\maketitle
\begin{abstract}
The increasing volume of scientific research necessitates effective communication across language barriers. Machine translation (MT) offers a promising solution for accessing international publications. However, the scientific domain presents unique challenges due to its specialized vocabulary and complex sentence structures. In this paper, we present the development of a collection of parallel and monolingual corpora for the scientific domain. The corpora target the language pairs Spanish-English, French-English, and Portuguese-English. For each language pair, we create a large general scientific corpus as well as four smaller corpora focused on the domains of: Cancer Research, Energy Research, Neuroscience, and Transportation research. To evaluate the quality of these corpora, we utilize them for fine-tuning general-purpose neural machine translation (NMT) systems. We provide details regarding the corpus creation process, the fine-tuning strategies employed, and we conclude with the evaluation results.
\end{abstract}

\section{Introduction}
The growth of scientific research across disciplines has intensified the need for efficient communication and international collaboration that transcends language barriers. While English is the dominant language of scientific publications~\cite{Altbach:2007}, a substantial volume of valuable academic work is produced in other languages. According to a 2019 study ~\cite{Stockemer2019}, almost 40\% of the articles of non-anglophone researchers are submitted in a language other than English. ~\cite{Altbach:2007}. Machine translation (MT) offers a compelling solution, providing access to a vast pool of international research and fostering seamless collaboration among researchers worldwide.

Advancements in Neural Machine Translation have significantly improved the quality of translations in various domains, largely driven by the application of the Transformer architecture~\cite{Vaswani:2017}. Transformers revolutionized NMT by enabling efficient parallel processing of entire sequences, leading to significant improvements in translation quality and fluency. However, the performance of NMT models often suffers when translating specialized domains due to the presence of specific terminology and sentence structures. Translating scientific text presents unique challenges distinct from general language translation~\cite{Byrne:2012}. Scientific domains are characterized by:

\begin{itemize}
\itemsep 0mm
\item {\bf Specialized Lexicon:} These domains employ a rich vocabulary of technical terms and abbreviations often absent from general language corpora.
\item {\bf Syntactic Complexity:} Scientific writing frequently utilizes complex sentence structures to convey precise relationships and subtle meanings.
\item {\bf Domain-Specific Discourse:} Each scientific domain possesses its own unique discourse patterns and conventions that are critical to understand for accurate translation.
\item {\bf Indeterminacy:} The division of scientific areas into numerous sub-disciplines and the existence of multidisciplinary works further complicate the definition of what a scientific domain encompasses.
\end{itemize}

These factors can significantly impair the performance of generic NMT models, leading to mistranslations or loss of scientific meaning.

To address these challenges, this paper presents the development of domain-specific corpora for scientific research and their application towards creating NMT models for four academic domains. We aimed at using the open-source NMT models which exhibit the best generic performance as base models and fine-tune them on these domain-specific corpora, so as to achieve translations that are not only fluent but also accurate and faithful to the scientific content. We focus on the Spanish--English (ES--EN), French--English (FR--EN), and Portuguese--English (PT--EN) translation directions, creating parallel corpora for each, alongside monolingual corpora for the respective languages (i.e., English, Spanish, French, and Portuguese). Our corpora encompass a large general scientific corpus and smaller sub-corpora dedicated to the research areas of:

\begin{itemize}
\itemsep 0mm
\item Cancer Research
\item Energy Research
\item Neuroscience
\item Transportation Research
\end{itemize}

We begin by presenting our dataset creation process which exploits the wealth of parallel titles and abstracts from bachelor and master theses, doctoral dissertations, and other scientific publications (such as published books and articles). Next, we outline the methodology of fine-tuning pre-trained NMT models using the aforementioned datasets. Finally, we evaluate the resulting models' performance by contrasting their output with the translations received from the original general-purpose NMT models as well as Google Translate. 

\section{Related Work}

Even though there is a plethora of parallel corpora in the ever-growing OPUS collection \cite{tiedemann-2012-parallel}, there is a shortage of those which aim at addressing parallel data acquisition for the scientific and academic domains, relatively to the importance and challenges of scientific translation. 

In OPUS, there are 2 parallel datasets of note, namely CAPES \cite{soares2018parallel} and SciELO \cite{soares2019large}. CAPES originates from the database of the Theses and Dissertations Catalog (TDC) and contains approximately 1.2 million sentence pairs for the EN--PT language pair \cite{soares2018parallel}, mined from theses and dissertation abstracts of students in post-graduate programs across Brazilian universities. Similarly, the SciELO parallel corpus has been extracted through the use of the SciELO database, which contains a broad range of open-access scientific articles. SciELO consists of approxiately 3.3 million parallel sentences and metadata for English, Spanish, and Portuguese, some of which are trilingually aligned (i.e., EN--ES--PT) \cite{soares2019large}. Both corpora are evaluated manually, as well as automatically by training and evaluating MT systems.

In order to address the parallel data gap for scientific texts among underrepresented European languages, the SciPar corpus \cite{roussis2022scipar} was created and made publicly availabe via the ELRC-SHARE repository.\footnote{\url{https://elrc-share.eu/}} SciPar contains 9.17 million sentence pairs in 31 language pairs and has been constructed from the titles and abstracts of bachelor and master theses, doctoral dissertations, and other scientific publications. It has been constructed through collecting, parsing, and processing metadata from 86 institutional repositories, digital libraries of universities, and national archives.

The \textit{Translations and Open Science} project focused on building EN--FR NMT systems for three pilot domains: (a) Climatology and Climate Change, (b) Neurosciences, and (c) Human Mobility, Environment, and Space. Fiorini et al.~\shortcite{fiorini2023translations} collected 316,701 parallel segments and 1,112 bilingual terms for these three domains, trained a generic NMT model from scratch, and the fine-tuned it on the collected datasets. The paper describes the manual and automatic evaluation and comparison that was conducted to determine domain-specific translation quality, while it also shows that adding data from SciPar results in further improvements.

\begin{table*}[ht]
\centering
\label{tab:domain-sentences}
\begin{tabular}{lccc}
\toprule
\textbf{Domain} & \textbf{EN--ES} & \textbf{EN--PT} & \textbf{EN--FR} \\
\midrule
Cancer & 57,226 & 123,357 & 49,112 \\
Energy & 107,710 & 205,662 & 87,918 \\
Neuroscience & 40,467 & 85,717 & 45,650 \\
Transp/tion & 26,795 & 35,181 & 19,151 \\
Gen. Scient. & 3,913,214 & 5,255,552 & 1,648,200 \\
\midrule
\textbf{Total} & 4,145,412 & 5,705,469 & 1,850,031 \\
\bottomrule
\end{tabular}
\caption{Parallel Corpora Sizes per Language Pair and Domain}
\end{table*}

\section{Dataset Creation}

In this section, we outline the end-to-end pipeline used for mining high-quality parallel and monolingual corpora from the titles and abstracts of theses and dissertations. First, we detail the strategy used to process approximately 9.3 million records from 62 academic repositories to extract 11,700,912 sentence pairs. A detailed list of the repositories can be seen at Table 4 of Appendix A. Then, we present the sizes of the parallel and monolingual corpora which resulted after applying various filtering methods, and, finally, we provide a brief documentation of the process used to create benchmark developer and test sets.

\begin{table*}[ht]
\centering
\label{tab:monolingual-sentences}
\begin{tabular}{lcccc}
\toprule
& \textbf{EN} & \textbf{ES} & \textbf{FR} & \textbf{PT} \\
\midrule
Cancer & 393,488 & 76,296 & 6,933 & 33,947 \\
Energy & 342,144 & 228,818 & 9,479 & 66,654 \\
Neuroscience & 262,618 & 47,112 & 5,380 & 19,640 \\
Transp/tion & 33,509 & 65,374 & 2,934 & 13,376 \\
Gen. Scient. & 13,187,215 & 10,512,255 & 753,487 & 3,335,615 \\
\midrule
\textbf{Total} & 14,218,974 & 10,929,855 & 778,213 & 3,469,232 \\
\bottomrule
\end{tabular}
\caption{Monolingual Corpora Sizes per Language Pair and Domain}
\end{table*}

\subsection{Repository Processing}

 Our strategy is directed toward structured metadata extraction and processing of academic records in order to mine domain-specific monolingual and parallel sentences, while also facilitating differences among various repositories. It constitutes a unified framework that builds upon and extends the approach used in SciPar \cite{roussis2022scipar}. The various steps used in our repository processing pipeline are the following:

\begin{itemize}
    \item \textbf{Acquisition of academic records:} Initially, we utilized the GNU Wget\footnote{\url{https://www.gnu.org/software/wget/}} package to automatically download all the records across the 62 repositories as HTML files.
    
    \item \textbf{Custom repository configuration:} In order to extract structured metadata across repositories which differ in the way they store information, we created a configuration file for each one (see Appendix B for an example). This required manual inspection of several HTML files from each repository and configure custom regex patterns and rules. Furthermore, we defined minimum character lengths for the extraction of abstracts and titles to ensure the validity of the extracted data.
    
    \item \textbf{Parsing and metadata extraction:} We utilized the Beautiful Soup\footnote{\url{https://www.crummy.com/software/BeautifulSoup/}} package and regex pattern processing to parse the records and extract structured metadata. This step used the custom configuration files for each repository and resulted in the construction of a JSON file for each record, organizing the data for further processing (see Appendix C for an example).
    
    \item \textbf{Domain classification:} Leveraging a simple keyword-based classification method, each record was categorized into one of four distinct domains: Cancer Research, Transportation Research, Energy Research, and Neuroscience. Records not fitting exclusively into these categories were classified as belonging to the "General Academic" domain.
    
    \item \textbf{Text extraction:} From each JSON's titles and abstracts, both monolingual and parallel documents were extracted. This process preserved information regarding the identified domain and language of each record. Candidate parallel documents were created when two or more titles and/or abstracts were present. The NLTK library\footnote{\url{https://www.nltk.org/}} was employed to split abstracts into sentences.
    
    \item \textbf{Parallel sentences mining:} Utilizing LASER and margin-based scoring \cite{artetxe2019margin,artetxe2019massively}, we extracted parallel sentences from the candidate parallel documents. We used 2 NVIDIA 2080 Ti GPUs for this process.
\end{itemize}

\subsection{Domain-specific Parallel Corpora}

In order to construct domain-specific parallel corpora for all the language pairs that we targeted, we concatenated the parallel sentences extracted earlier from each one of the 62 repositories into domain and language pair specific files. 

Additionally, a LASER alignment score threshold of 0.98 was applied which, although being lower than thresholds used in other works which mine parallel sentences in a global way, ensures that parallel data originating from titles are not discarded \cite{roussis2022scipar}.

Finally, we deduplicated all parallel corpora and filtered them by removing sentence pairs which: (a) have identical source and target sentences, (b) contain an empty sentence in either side, (c) consist of more than 250 words in either sentence, (d) are solely comprised of digits, (e) are identified as belonging to incorrect languages, or {f} consist mostly of URLs and e-mails \cite{papavassiliou2018ilsp}.

The resulting domain-specific parallel corpora in Table 1 sum up to 11,700,912 sentence pairs in total. We can see that the the "General Scientific" domain considerably surpasses all other domains combined in corpus size across all language pairs. Furthermore, the corpus sizes for the EN--PT and EN--ES language pairs generally exceed those for the EN--FR language pair, which is indicative of the plethora of academic repositories originating from Latin American countries. Among the four focused domains, the "Energy" and "Cancer Research" domains exhibit higher data representations. The disparities in corpus sizes across different domains reflect the extent of each domain's scope, and the inherent challenges of domain classification.

In Table 2, we list the sizes of the domain-specific monolingual corpora for each of the four targeted languages. These results do not include the monolingual sentences in the parallel data and have been extracted from academic records without parallel titles and/or abstracts. It can be observed that there are many variations among the domains that we focused on; The "General Scientific" domain is significantly larger that all the other domains that we focused on. However, more pronounced differences concern the data availability for each language, as, for example, the English monolingual corpus is two orders of magnitude larger than the French one. Although we did not directly used monolingual data in this work, they can potentially enhance the performance of NMT models with the use of back-translation \cite{sennrich2015improving,edunov2018understanding}, as well as constitute a high-quality part of a pre-training dataset for a Large Language Model (LLM).

\subsection{Benchmark Creation}

The benchmark creation process was tailored to meet the specific demands of evaluating our fine-tuned models across the four scientific domains that we focus on. Our goal was to create:
\begin{itemize}
\itemsep 0mm
\item \textbf{Developer set}s which can be used to monitor the training process and select optimal hyperparameters
\item \textbf{Test sets} which can be used to evaluate the generalization capabilities of our models, as well as to compare them with the base OPUS-MT models and Google Translate.
\end{itemize}

For each of the four targeted domains, 1,000/1,000 sentence pairs were designated for development/testing, whereas the "General Scientific" set was allocated 3,000/3,000 sentence pairs. In order to build the domain-specific developer and test sets, we followed a structured multi-stage process. We started from the filtered parallel corpora for each domain and language pair, and applied a LASER threshold greater than 1.08 to identify sentences with substantial semantic alignment. Additionally, we used a stricter token ratio filter (1.66) and removed sentence pairs in which either sentence is less than 3 words long. 

Afterwards, we sample 2,000 unique records for each domain-specific dataset and 6,000 for the "General Scientific" datasets, from which a single sentence pair was randomly selected for inclusion in the combined developer/test sets. In other words, each parallel sentence in the sets is derived from a single thesis/abstract from a pool of randomly selected ones. 

For the construction of developer and test sets, we equally distributed the sentence pairs into two, resulting in 1,000 parallel sentences for each domain-specific developer and test set and 3,000 parallel sentences for the "General Scientific" developer and test set.

\section{Domain Adaptation for NMT}

Using domain adaptation, we can enhance the ability of an existing MT system to translate documents from a very specific domain, through the utilization of an in-domain parallel corpus. To this end, we chose to use selected pre-trained models from OPUS-MT \cite{tiedemann-2020-tatoeba} as our baseline systems. We chose these models because they are all based on state-of-the-art transformer-based neural machine translation architectures, and they were trained on freely available parallel corpora from the OPUS3 bitext repository. Even though these models are not considered to produce the best quality for the selected language pairs, they provide a robust initial performance across a variety of language pairs. Additionally, they have also been trained on general domain texts, making them ideal for our purpose, as they have seen little to no data from the scientific domain.

Since we decided to work on French, Spanish and Portuguese, we selected the open-source OPUS-MT models \cite{tiedemann-thottingal-2020-opus} for the FR--EN, ES--EN, and PT--EN, translation directions as our base models. We aimed at using the largest available OPUS-MT models for each language pair and made use of the latest ones which where trained for the Tatoeba Challenge \cite{tiedemann-2020-tatoeba}, an open project encouraging people to develop machine translation in real-world cases for many languages. In particular, we used the following models:
\begin{itemize}
    \item \textbf{FR--EN:} We selected the transformer-big variant, trained on English--French parallel data.\footnote{opusTCv20210807+bt\_transformer-big\_2022-03-09 \par \url{https://github.com/Helsinki-NLP/Tatoeba-Challenge/tree/master/models/fra-eng}}. We use a learning rate of 7e-5 and 0.2 dropout for all the fine-tuning experiments.

    \item \textbf{ES--EN:} We chose to use the transformer-big variant which was trained for translating Catalan (CA), Occitan (OC), and Spanish (ES) texts into English,\footnote{opusTCv20210807+bt\_transformer-big\_2022-03-13 \par \url{https://github.com/Helsinki-NLP/Tatoeba-Challenge/tree/master/models/cat+oci+spa-eng}} as a larger variant was not available for the ES--EN translation direction. We use a learning rate of 7e-5 and 0.2 dropout for all the fine-tuning experiments.
    
    \item \textbf{PT--EN:} For the PT--EN translation direction, we used the transformer-align \cite{bahdanau2014neural} variant,\footnote{opus+bt-2021-04-30 \par \url{https://github.com/Helsinki-NLP/Tatoeba-Challenge/tree/master/models/por-eng}} as there were not any variants with more parameters. We use a learning rate of 5e-5 and 0.1 dropout for all the fine-tuning experiments.
\end{itemize}

\begin{table*}[htbp]
\centering
\label{tab:domain_adaptation}
\begin{tabular}{lcccccccccccccc}
\toprule
& \multicolumn{3}{c}{Average of 4 domains} & \multicolumn{3}{c}{General Scientific} \\
\cmidrule(lr){2-4} \cmidrule(lr){5-7}
 & BLEU & chrF2++ & COMET & BLEU & chrF2++ & COMET \\
\midrule
\textbf{ES--EN} \\
Base OPUS-MT & 49.7 & 70.5 & 69.5 & 51 & 71.7 & 68.9 \\
+ FT w/ IND & 50.7 & 71.3 & 70.4 & 52.1 & 72.2 & 68.9 \\
+ FT w/ IND \& GSC & \textbf{51.9} & \textbf{71.7} & 70.9 & \textbf{54} & \textbf{73.1} & 71 \\
Google Translate & 51 & 71.1 & \textbf{72.2} & 52.6 & 72.4 & \textbf{73} \\
\midrule
$\Delta$(Ours – Baseline) & +2.2 & +1.2 & +1.4 & +3 & +1.4 & +2.1 \\
\midrule
\textbf{PT--EN} \\
Base OPUS-MT & 46 & 68.3 & 66.7 & 44.9 & 67.7 & 66.3 \\
+ FT w/ IND & 48.1 & 69.7 & 66.8 & 46.6 & 68.7 & 66.1 \\
+ FT w/ IND \& GSC & \textbf{48.4} & \textbf{69.9} & 67.6 & \textbf{47.3} & \textbf{69.1} & 67.8 \\
    Google Translate & 48 & 69.3 & \textbf{70.6} & 46.7 & 68.6 & \textbf{70.5} \\
\midrule
$\Delta$(Ours – Baseline) & +2.4 & +1.6 & +0.9 & +2.4 & +1.4 & +1.5 \\
\midrule
\textbf{FR--EN} \\
Base OPUS-MT & 37.6 & 63.6 & 57.5 & 38.4 & 63.3 & 57.2 \\
+ FT w/ IND & 39.1 & 65 & 58.9 & 39.4 & 64.5 & 58.4 \\
+ FT w/ IND \& GSC & 40.2 & \textbf{65.7} & 59.9 & 40.7 & \textbf{65.3} & 59.5 \\
Google Translate & \textbf{41} & 65 & \textbf{62.2} & \textbf{40.9} & 64.4 & \textbf{61.6} \\
\midrule
$\Delta$(Ours – Baseline) & +2.6 & +2.1 & +2.4 & +2.3 & +2 & +2.3 \\
\bottomrule
\end{tabular}
\caption{Results of Domain Adaptation Experiments}
\end{table*}

All of these variants have been originally trained on subsets of the OPUS-MT corpus, along with additional back-translated data; a method which has been widely adopted by the research community to further improve translation quality \cite{sennrich2015improving,edunov2018understanding}. For fine-tuning the OPUS-MT models we used the MarianMT framework  \cite{junczys2018marian}, a decision driven by MarianMT’s efficiency, flexibility, and open-source nature, making it particularly suited for research purposes.

Fine-tuning was conducted using the curated corpora which resulted from the filtering process described in section 3.2, comprising parallel sentences from academic texts across our target domains: Cancer Research, Energy Research, Neuroscience, and Transportation Research. These corpora were further enriched with general academic texts to provide a broader linguistic context, thereby mitigating the potential for overfitting to domain-specific jargon and syntax. In our domain adaptation experiments, we make use of all developer sets described in Section 3.3 (7000 sentence pairs in total) so that the fine-tuning process does not continue if the BLEU or chrF scores do not improve for 5 consecutive checkpoints (every 500 steps). Batch size is automatically determined by MarianNMT to fit reserved GPU memory, while parameters are updated every 2 batches \cite{ott2018scaling}.

During the fine-tuning phase, particular attention was paid to the balance between domain-specific and general academic texts, ensuring that the models retained their generalizability while enhancing their domain-specific performance. Model hyperparameters were meticulously optimized through a series of experiments, with evaluation metrics such as BLEU \cite{papineni2002bleu}, chrF2++ \cite{popovic2015chrf,popovic2016chrf,popovic2017chrf++}, and COMET \cite{rei2020comet} used to monitor improvements in translation quality.

\section{Results \& Evaluation}

In order to determine the improvement on translation quality for each specific domain of interest, we perform evaluation using all the 4 domain-specific test sets, as well as the General Scientific test set (see Section 3.3). To that end, the most widely-used benchmarks for the evaluation of automatic translations have been chosen, these being BLEU, chrF2++, and COMET. BLEU measures how much overlap there is between the translated text and a reference translation in terms of n-gram phrases, while chrF2++ considers character-level accuracy and morphology in addition to n-grams. Unlike BLEU and chrF2++ which focus on word-level overlap, COMET doesn't directly compare the words themselves. Instead, it uses the XLM-R Large as its backbone and estimates how well the translated text captures the meaning conveyed in the reference text. COMET, in particular, has been shown to be superior to the other two metrics that we used, as it better correlates with human judgements and provides a robust way to differentiate high-performing systems \cite{rei2020comet}.

We use the SacreBLEU toolkit \cite{post2018call} for evaluation of BLEU\footnote{nrefs:1\textbar{} bs:1000\textbar{} seed:12345\textbar{} case:mixed\textbar{} eff:no\textbar{} tok:13a\textbar{} smooth:exp\textbar{} version:2.0.0} and chrF2++\footnote{nrefs:1\textbar{} bs:1000\textbar{} seed:12345\textbar{} case:mixed\textbar{} eff:yes\textbar{} nc:6\textbar{} nw:2\textbar{} space:no\textbar{} version:2.0.0} as it ensures that they can be computed in a reproducible way, while the wmt20-comet-da\footnote{\url{https://huggingface.co/Unbabel/wmt20-comet-da}} version was used for computing the COMET score.

In Table 2, we can see that our fine-tuned models outperform the base models. In all language pairs and across all metrics, the systems that use fine-tuning with in-domain data (either alone or combined with general scientific text) outperform the baseline OPUS-MT system by +2.4 BLEU and +1.6 COMET on average. Fine-tuning the baseline OPUS-MT models with in-domain parallel data (+FT w/IND) yields an average improvement of +1.5 BLEU and +0.8 COMET. 

Across all language pairs, adding general scientific text for fine-tuning on top of in-domain data (+ FT w/ IND \& GSC) leads to further improvements compared to just using in-domain data, improving by +0.9 BLEU and +0.8 COMET. This suggests that general scientific data can act as a helpful supplement for domain adaptation, although it should be noted that their number is an order of magnitude larger than the in-domain data.

While the fine-tuned systems achieve the best scores in the majority of cases, Google Translate remains highly competitive, producing the highest score in COMET and outperforming all other systems in almost all metrics for FR--EN. This indicates that even without specific domain adaptation, Google Translate offers strong translation capabilities for high-resource language pairs, even for niche domains like the ones selected for this experiment.  

\section{Conclusion}

Our work presents the development of domain-specific corpora for scientific research and their application towards creating improved NMT models for four scientific domains: Cancer Research, Energy Research, Neuroscience, and Transportation Research. We acquired 11.7M parallel sentences for three language pairs (EN--ES, EN--PT, and EN--FR) by processing the records from 62 academic repositories. Our aim was to leverage the strengths of existing open-source NMT models while specializing them in these particular domains through fine-tuning.

Our findings indicate that fine-tuning generic NMT models with domain-specific parallel data leads to substantial improvements in translation quality for the targeted scientific domains. Additionally, by including general scientific text alongside domain-specific data during fine-tuning offers further enhancements, potentially by providing broader linguistic context.

This work contributes to the ongoing effort to bridge the language gap in scientific research by developing NMT models that can accurately and fluently translate scientific text across various domains.

Looking ahead, there are a lot of ideas that we could implement in order to further improve our results.  Multi-domain adaptation holds promise for creating models adaptable to a wider range of scientific content. Backtranslation, especially with domain-specific tagging and iterative approaches, offers another potential path for significant improvements. Additionally, advanced parallel sentence filtering techniques, like those offered by Bicleaner AI, can ensure high-quality training data. By incorporating these advancements, we can contribute to NMT models that effectively bridge the communication gap in the global scientific community.

\section*{Acknowledgements}

This work was created within the Scilake\footnote{\url{https://scilake.eu/}} project. We are grateful to the Scilake project for providing the resources and support that made this work possible. This project has received funding from the European Union’s Horizon Europe framework programme under grant agreement No. 101058573.


\bibliography{eamt24}
\bibliographystyle{eamt24}

\clearpage
\appendix
\section*{Appendix A}
\subsection*{List of the 62 Processed Repositories}

\scriptsize
\begin{table*}[htbp]
\centering
\caption{Processed Repositories Names and URLs}
\label{tab:processed-repositories}
\small
\begin{tabular}{p{0.22\textwidth} p{0.26\textwidth} | p{0.22\textwidth} p{0.26\textwidth}}
\toprule
\textbf{Name of Repository} & \textbf{URL} & \textbf{Name of Repository} & \textbf{URL} \\
\midrule
ESTUDO GERAL - Coimbra & \url{https://estudogeral.uc.pt/} & Repositorio U. de Lisboa & \url{https://repositorio.ul.pt/} \\
RepositóriUM - U. Minho & \url{https://repositorium.sdum.uminho.pt/} & Repositório Lusófona & \url{https://recil.ensinolusofona.pt/} \\
Repositório U. da Madeira & \url{https://digituma.uma.pt/} & RIA - U. de Aveiro & \url{https://ria.ua.pt/} \\
RUN - U. Nova & \url{https://run.unl.pt/} & Repositório U. Fernando Pessoa & \url{https://bdigital.ufp.pt/} \\
Repositorio Aberto - U. Aberta & \url{https://repositorioaberto.uab.pt/} & Repositorio de UTAD & \url{https://repositorio.utad.pt/} \\
Biblioteca Digital do IPB & \url{https://bibliotecadigital.ipb.pt/} & Dadun - U. of Navarra & \url{https://dadun.unav.edu/} \\
UIBrepository - U. Illes Balears & \url{https://dspace.uib.es/} & Archivo Digital - UPM & \url{https://oa.upm.es/} \\
Dipòsit Digital de la UB & \url{http://diposit.ub.edu/} & RODERIC - U. de Valencia & \url{https://roderic.uv.es/} \\
UCrea - U. de Cantabria & \url{https://repositorio.unican.es/} & GREDOS - U. de Salamanca & \url{https://gredos.usal.es/} \\
Repository of UOC & \url{https://openaccess.uoc.edu/} & RIULL - U. de La Laguna & \url{https://riull.ull.es/} \\
idUS - U. de Sevilla & \url{https://idus.us.es/} & UVaDOC - U. de Valladolid & \url{https://uvadoc.uva.es/} \\
UPF Digital Rep. (Barcelona) & \url{https://repositori.upf.edu/} & RiuNet - U. de Valencia - UPV & \url{https://riunet.upv.es/} \\
RUC - U. de Coruna & \url{https://ruc.udc.es/} & UPCommons - UPC & \url{https://upcommons.upc.edu/} \\
UAM - U. Autónoma de Madrid & \url{https://repositorio.uam.es/} & DUGiDocs - U. de Girona & \url{https://dugi-doc.udg.edu/} \\
Repositori U. Jaume I & \url{https://repositori.uji.es/} & AccedaCRIS - ULPGC & \url{https://accedacris.ulpgc.es/} \\
DDD - U. Autònoma de Barcelona & \url{https://ddd.uab.cat/} & Tesis Doctorals en Xarxa & \url{https://www.tdx.cat/} \\
Repositorio U. de Chile & \url{https://repositorio.uchile.cl/} & Bib. Institucional - U. Andres Bello & \url{https://repositorio.unab.cl/} \\
Repositorio Institucional UCA & \url{https://repositorio.uca.edu.ar/} & DSpace PUCP (Perú) & \url{https://tesis.pucp.edu.pe/} \\
Repositorio U. Nac. de Ingeniería & \url{http://cybertesis.uni.edu.pe/} & Repositorio Académico UPC & \url{https://repositorioacademico.upc.edu.pe/} \\
Repositorio U. de los Andes & \url{https://repositorio.uniandes.edu.co/} & Repositorio U. Nac. de Colombia & \url{https://repositorio.unal.edu.co/} \\
Repositorio U. Javeriana & \url{https://repository.javeriana.edu.co/} & U. de Brasilia - Institutional Rep. & \url{https://repositorio.unb.br/} \\
UFSC - Institutional Repository & \url{https://repositorio.ufsc.br/} & UFPB - Institutional Repository & \url{https://repositorio.ufpb.br/} \\
RIUT - UTFPR & \url{http://repositorio.utfpr.edu.br/} & Repositório Institucional UNESP & \url{https://repositorio.unesp.br/} \\
LUME - UFRGS & \url{https://lume.ufrgs.br/} & Repositorio UFRN & \url{https://repositorio.ufrn.br/} \\
Repositório Institucional UNIFESP & \url{https://repositorio.unifesp.br/} & ATTENA - Digital da UFPE & \url{https://repositorio.ufpe.br/} \\
Repositorio UFBA & \url{https://repositorio.ufba.br/} & Inst. Repository PUCRS & \url{https://repositorio.pucrs.br/} \\
Repositorio UFU & \url{https://repositorio.ufu.br/} & Locus Repositório UFV & \url{https://www.locus.ufv.br/} \\
DIAL - U. Catholique de Louvain & \url{https://dial.uclouvain.be/pr/boreal/} & Papyrus - U. de Montreal & \url{https://papyrus.bib.umontreal.ca/} \\
Savoirs UdeS - U. de Sherbrooke & \url{https://savoirs.usherbrooke.ca/} & Toulouse Capitole Publications & \url{https://publications.ut-capitole.fr/} \\
RED Repositorios Latinoamericanos & \url{https://repositorioslatinoamericanos.uchile.cl/} & Revistas Academicas Chilenas & \url{https://revistaschilenas.uchile.cl/} \\
HAL Theses - Theses en Ligne & \url{https://theses.hal.science/} & HAL Open Science & \url{https://hal.science/} \\
\bottomrule
\end{tabular}
\end{table*}
\normalsize

\clearpage
\onecolumn  

\section*{Appendix B}

\subsection*{Example Configuration File for Processing the HTMLs}

The configuration files used for each one of the repositories that we processed facilitate the structured extraction of metadata. The regex patterns specified within the configuration file are used to match the HTML structure of the academic records, while the minimum lengths specify the minimum characters that a title or abstract must have so as to be considered valid. Additionally, the file specifies the targeted languages for which we extract texts. 

Below is the JSON configuration file for the DIAL UCLouvain repository:

\footnotesize
\begin{verbatim}
{
    "abstracts_regex": ".*publication-metadata.*",
    "abstracts_min_len": 20,
    "titles_regex": ".*citation_title.*",
    "titles_min_len": 20,
    "keywords_regex": ".*Keywords.*",
    "authors_regex": ".*citation_author.*",
    "publishers_regex": ".*Affiliation.*|.*Publisher.*",
    "date_available_regex": ".*Publication date.*|.*Defense date.*",
    "journal_regex": ".*citation_journal_title.*|.*citation_dissertation_institution.*",
    "bibliographic_citation_regex": ".*Bibliographic reference.*",
    "document_language_regex": ".*Language.*",
    "link_html_regex": ".*Permanent URL.*",
    "link_pdf_regex": ".*citation_pdf_url.*",
    "document_type_regex": ".*Document type.*",
    "license_regex": ".*Access type.*",
    "URI_regex": ".*Permanent URL.*",
    "targeted_langs": [
        "en",
        "es",
        "pt",
        "fr"
    ]
}
\end{verbatim}
\normalsize

\clearpage
\section*{Appendix C}
\subsection*{Example JSON File Extracted from a Record}

Our repository processing pipeling results into JSON files with the following structured metadata: abstracts,  titles, repository name, links to the HTML and PDF versions of the documents, URI, license type, publication dates, document language and type (e.g., thesis), keywords, author names, publisher information, publication dates, journal titles, bibliographic citations, and the identified domains.

Below is an example JSON file extracted from a record, classified as belonging to the "Energy Research" domain, which originates from the Biblioteca Digital do IPB - Instituto Politecnico de Braganca repository:

\begin{sloppypar}
\footnotesize
\begin{verbatim}
{
    "abstracts": {
        "en": "In this research is intended to analyse the expansion of the 
               economic sector related to the development ways of renewable 
               energy and the economic and financial performance of companies
               operating in this field. [...]",
        "pt": "Esta investigação pretende analisar a expansão do 
               setor económico relacionado com o desenvolvimento das 
               energias renováveis e os desempenhos económico e financeiro 
               das empresas que operam nesse setor. [...]"
    },
    "titles": {
        "en": "The development ways of renewable energy: the economic and financial 
               performance of firms in this sector in Armenia and OECD countries"
    },
    "repository": "bibliotecadigital-ipb-pt",
    "html_id": 14638,
    "link_html": "https://bibliotecadigital.ipb.pt/handle/10198/14638",
    "link_pdf": "https://bibliotecadigital.ipb.pt/bitstream/
                 10198/14638/1/Tarakhchyan_Siranush.pdf",
    "uri": "http://hdl.handle.net/10198/14638",
    "license_link": "http://creativecommons.org/licenses/by-nc/4.0/",
    "license": "openAccess",
    "date_available": "2017-11-20T15:08:42Z",
    "document_language": "en",
    "document_type": "masterThesis",
    "keywords": [
        "Renewable energy (RE)",
        "Financial data",
        "Financial ratios",
        "Market price",
        "Environment",
        "Domínio/Área Científica::Ciências Sociais::Economia e Gestão"
    ],
    "authors": [
        "Tarakhchyan, Siranush"
    ],
    "publishers": [],
    "bibliographic_citation": "",
    "journal": "",
    "domain_keyword_count": {
        "cancer": 0,
        "energy": 6,
        "transportation": 0,
        "neuroscience": 0
    },
    "domain": "energy"
}
\end{verbatim}
\end{sloppypar}
\normalsize

\end{document}